\documentclass[runningheads]{llncs}

\usepackage[review,year=2026,ID=436]{eccv}

\usepackage{eccvabbrv}

\usepackage{graphicx}
\usepackage{booktabs}
\usepackage{tikz}
\usepackage[accsupp]{axessibility}  
\usepackage{booktabs} 
\usepackage{multirow} 
\usepackage{wrapfig}

\usepackage{hyperref}

\makeatletter
\def\blfootnote{\gdef\@thefnmark{}\@footnotetext}
\makeatother

\usepackage{multibib}
\newcites{supp}{Supplementary References}

\usepackage{orcidlink}
\usepackage{marvosym}

\usepackage{multibib}
\newcites{supp}{Supplementary References}

\begin{document}
\nolinenumbers
\title{Real-Time Human Frontal View Synthesis from a Single Image} 

\titlerunning{PrismMirror}

\author{Fangyu Lin\inst{1}\textsuperscript{$\star$} \and
Yingdong Hu\inst{1}\textsuperscript{$\star$} \and
Lunjie Zhu\inst{1} \and
Zhening Liu\inst{1} \and
Yushi Huang\inst{1} \and
Zehong Lin\inst{1}\textsuperscript{$\dagger$} \and
Jun Zhang\inst{1}\textsuperscript{$\dagger$}}

\authorrunning{F. Lin~Y. Hu et al.}

\institute{HKUST, Hong Kong SAR, China \\
\email{\{flinao, yhudj\}@connect.ust.hk} \\
\email{\{eezhlin, eejzhang\}@ust.hk} \\
\url{https://github.com/rslinfy/PrismMirror}}

\maketitle

\blfootnote{\textsuperscript{$\star$} Equal contribution.}
\blfootnote{\textsuperscript{$\dagger$} Corresponding authors.}

\begin{abstract}
  Photorealistic human novel view synthesis from a single image is crucial for democratizing immersive 3D telepresence, eliminating the need for complex multi-camera setups. However, current rendering-centric methods prioritize visual fidelity over explicit geometric understanding and struggle with intricate regions like faces and hands, leading to temporal instability. Meanwhile, human-centric frameworks suffer from memory bottlenecks since they typically rely on an auxiliary model to provide informative structural priors for geometric modeling, which limits real-time performance. To address these challenges, we propose PrismMirror, a geometry-guided framework for instant frontal view synthesis from a single image. By avoiding external geometric modeling and focusing on frontal view synthesis, our model optimizes visual integrity for telepresence. Specifically, PrismMirror introduces a novel cascade learning strategy that enables coarse-to-fine geometric feature learning. It first directly learns coarse geometric features, such as SMPL-X meshes and point clouds, and then refines textures through rendering supervision. To achieve real-time efficiency, we distill this unified framework into a lightweight linear attention model. Notably, PrismMirror is the first monocular human frontal view synthesis model that achieves real-time inference at 24 FPS, significantly outperforming previous methods in both visual authenticity and structural accuracy. 
\end{abstract}

\vspace{-10mm}

\begin{tikzpicture}[remember picture, overlay]
    \node[
        anchor=north,      
        yshift=-0.05cm        
    ] at (current page.north) { 
        \includegraphics[width=2.5cm]{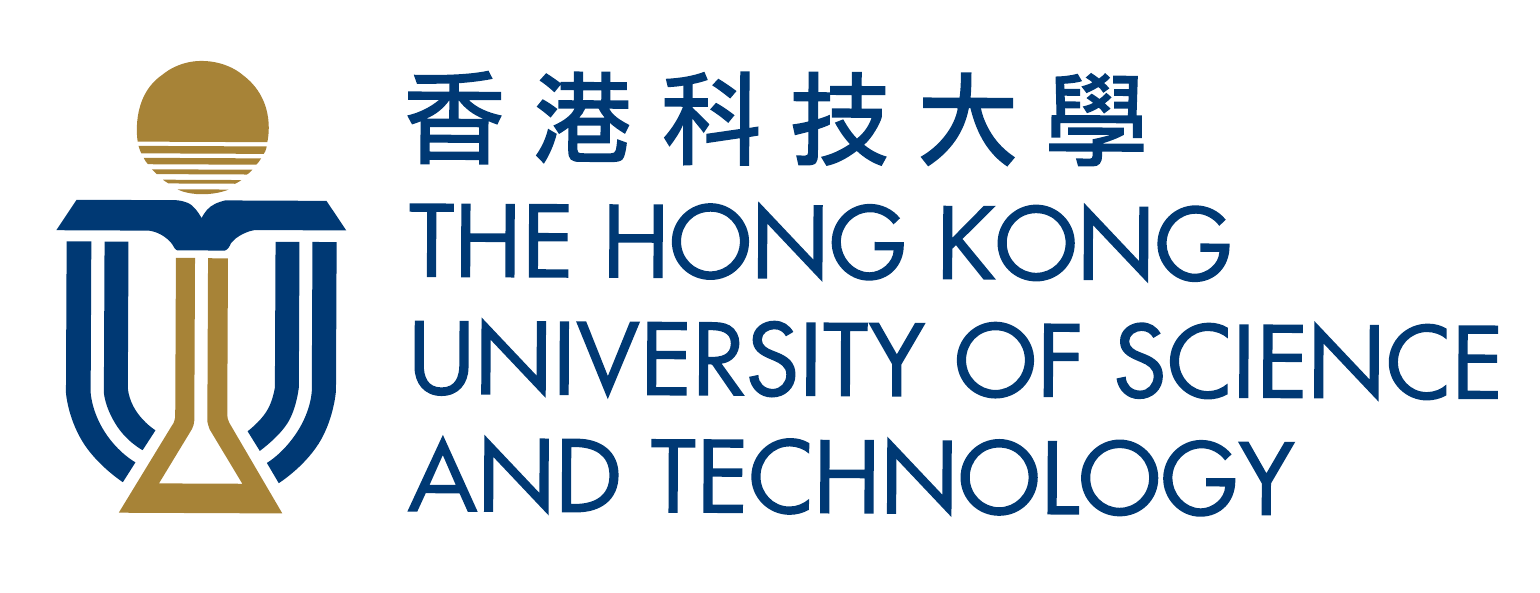}%
        \hspace{0.1cm}
        \includegraphics[width=3.5cm]{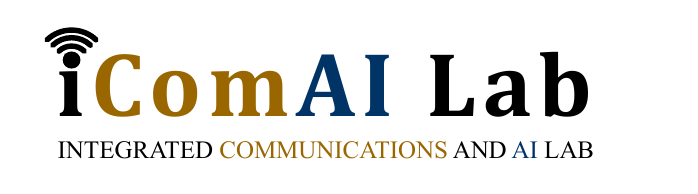}
    };
\end{tikzpicture}

\section{Introduction}
High-fidelity human novel view synthesis (NVS) from a single image represents a critical frontier in computer vision, particularly for enabling next-generation immersive 3D telepresence and virtual video conferencing. While photorealistic rendering has been achieved in controlled environments using sophisticated volumetric scanners and multi-camera setups, their reliance on expensive hardware and high latency impedes widespread consumer adoption. Consequently, the field has gravitated towards instant single-view reconstruction, aiming to synthesize realistic novel viewpoints from monocular inputs, such as a standard webcam feed. However, reconstructing a 3D volume from a 2D projection is an inherently ill-posed problem, plagued by depth ambiguity and structural inconsistencies.

\begin{figure}[t!]
  \centering
  \includegraphics[scale=0.4]{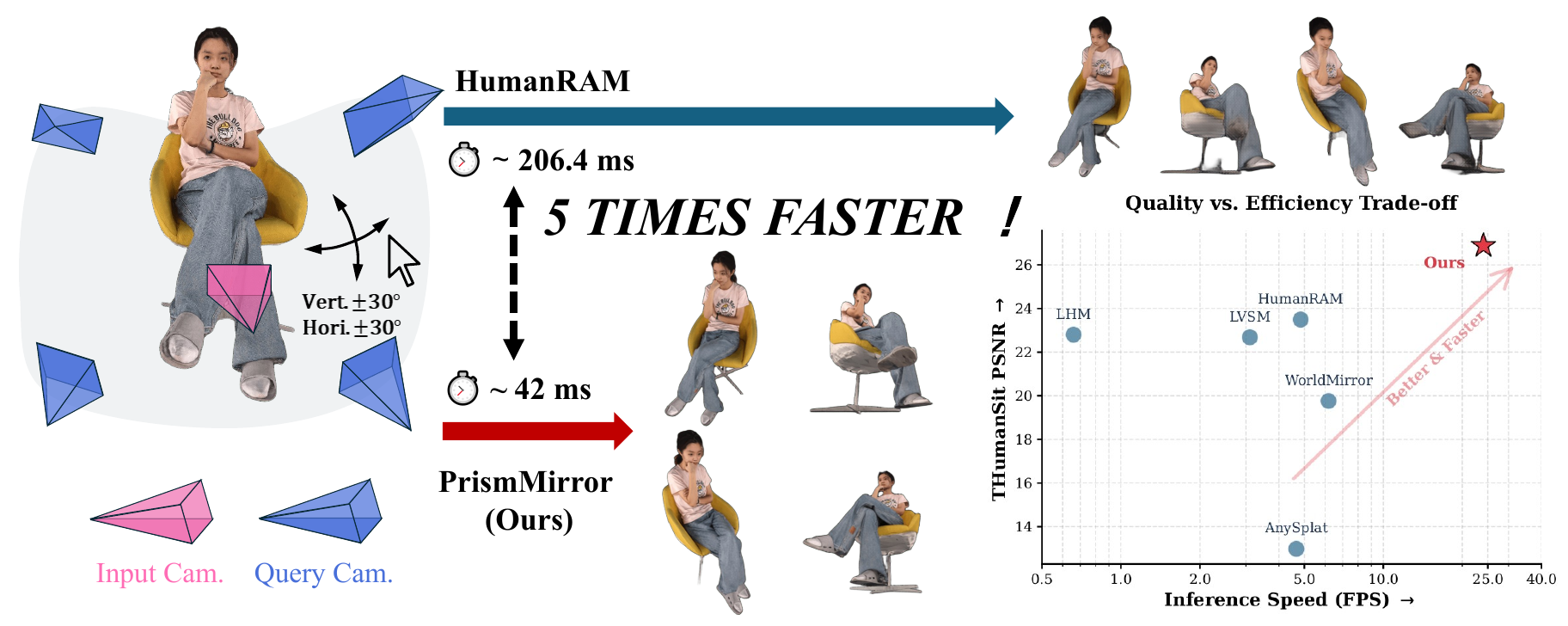} 
  \vspace{-1.4em}
  \caption{\textbf{Real-time high-fidelity human frontal view synthesis.} PrismMirror achieves a $5\times$ inference speedup ($\sim$42 ms vs. 206.4 ms) over HumanRAM, while maintaining superior visual quality and reaching real-time frame rates of 24+ FPS.}
  \label{fig:teaser}
  \vspace{-2em}
\end{figure}

Existing approaches to single-image NVS primarily adopt rendering-centric approaches, such as 3D Gaussian splatting (3DGS)~\cite{3dgs,dass,ffd3dgsc,mega} and neural radiance fields (NeRFs)~\cite{nerf}. These methods primarily focus on optimizing pixel-wise visual accuracy rather than explicit 3D geometric modeling, which often leads to spatial inconsistencies and unstable geometry. In live telepresence applications, these issues manifest as severe structural artifacts, including floating densities, distracting temporal jitter, and pronounced depth stretching. Meanwhile, methods based on generative diffusion models, while capable of producing high-fidelity novel views, rely on iterative denoising processes that introduce significant latency, making them unsuitable for real-time applications.

On the other hand, human-centric NVS poses challenges that are fundamentally distinct from and significantly more demanding than general scene reconstruction. Recent advancements in large view synthesis models (LVSMs) have attempted to address general 3D environments, with frameworks like efficient-LVSM~\cite{efficient-lvsm} accelerating inference through computationally efficient transformer architectures. Meanwhile, large reconstruction models like WorldMirror~\cite{worldmirror} and AnySplat~\cite{anysplat} have emerged to reconstruct multiple representations together into their pipelines. However, these approaches fail to meet the real-time performance demands of live telepresence. More importantly, generic scene models lack the capacity to capture the intricate biomechanical constraints and high-frequency details of human subjects. Tasks such as estimating parameters for expressive models like SMPL-X~\cite{smplx} require precise extraction of structural features, far exceeding the complexity of standard depth estimation. Moreover, human subjects contain semantically sensitive regions, such as facial micro-expressions and hand articulations, where even minor errors in texture alignment or geometry lead to perceptible identity drift. Consequently, generic models cannot reliably deliver the fidelity required for live human telepresence.

To address these limitations, recent efforts have integrated explicit human priors into the reconstruction pipeline. A notable example is HumanRAM~\cite{humanram}, which utilizes SMPL-X priors to condition transformer architectures for achieving animatable NVS. However, this method depends heavily on auxiliary models to provide structural priors, such as rendered position maps, and operates within a computationally expensive triplane feature space throughout its pipeline. These architectural choices lead to significant GPU memory overhead, hindering inference speed and falling short of the real-time responsiveness required for telepresence applications. Moreover, the reliance on external models and priors during inference compromises computational efficiency, which could be better allocated to high-fidelity texture rendering and real-time performance.

We argue that achieving real-time high-fidelity human NVS requires unifying geometric modeling directly within the framework itself. This enables the model to independently extract and utilize robust geometric features, eliminating the need for auxiliary models during inference and, consequently, reducing memory overhead while improving computational efficiency. To this end, we propose PrismMirror, a novel geometry-guided framework for instant single-image frontal view synthesis. By integrating explicit structural feature learning and refining these features within a targeted frontal view synthesis range, our approach achieves both high visual fidelity and real-time efficiency. Recognizing the practical demands of telepresence, which typically do not require full 360-degree view synthesis, we strategically constrain our model to a frontal viewing frustum of ±30 degrees. This design effectively alleviates the inherent challenges of hallucinating heavily occluded regions while allowing the model to focus on maintaining visual integrity in the frontal view, ensuring smooth and stable view transitions during telepresence sessions.

At the core of PrismMirror lies our novel \textit{cascade learning} strategy, which addresses the traditional tradeoff between structural accuracy and real-time inference speed, as shown in Fig.~\ref{fig:teaser}. The \textit{cascade} mechanism enforces a coarse-to-fine learning paradigm. In the front-end stage, the model learns foundational geometric features, such as SMPL-X meshes and point clouds, directly within the framework through strict supervision. This approach avoids reliance on memory-intensive auxiliary models during inference. In the back-end stage, the network focuses on refining fine-grained textures, employing view synthesis and 3DGS losses to generate photorealistic outputs with enhanced detail. To achieve real-time performance, we introduce a \textit{distillation} phase that compresses the learned geometric and texture features into an efficient backbone. This phase replaces computationally expensive FlashAttention mechanisms \cite{flashattention,flashattention2}  with lightweight linear attention \cite{fla,sana}, significantly reducing inference latency while preserving structural fidelity. By integrating these components into a unified framework, PrismMirror achieves state-of-the-art performance in both visual authenticity and structural accuracy while meeting the strict real-time constraints of telepresence applications. In summary, our contributions are as follows:

\begin{itemize}
    \item We propose PrismMirror, the first monocular human frontal view synthesis model capable of delivering authentic identity-preserving frontal human reconstruction with real-time performance of 24 FPS.
    \item We introduce a cascade distillation strategy that effectively decouples foundational geometric learning (e.g., SMPL-X meshes and point clouds) from fine-grained texture synthesis (NVS, 3DGS), leveraging a distillation process that leverages lightweight linear attention for exceptional efficiency.
    \item By enabling the model to directly extract and utilize geometric features without relying on auxiliary models, and by focusing on a practical frontal viewing range, PrismMirror overcomes the memory and computational bottlenecks of prior human-centric frameworks, achieving state-of-the-art visual fidelity and structural accuracy under real-time constraints.
\end{itemize}

\section{Related Works}

\subsection{Human Novel View Synthesis}
The reconstruction of novel views from sparse human observations has transitioned from implicit function learning to real-time volumetric rendering. Early pixel-aligned implicit approaches~\cite{pifu, pifuhd, thuman2, bodynet, deephuman, icon, econ} employ spatial features to recover geometry but struggle with resolution limitations and slow inference. Conversely, NeRFs~\cite{nerf, nhp, keypointnerf, sherf, pva} and 3DGS~\cite{3dgs,hugs,ghg,gst} have advanced the field by enabling photo-realistic rendering. Recent feed-forward methods~\cite{gpsgaussian,evagaussian,gpsgaussian+} attempt to directly regress 3D attributes, but they prioritize visual accuracy and treat geometry merely as an implicit byproduct. This lack of explicit geometric modeling causes spatial inconsistencies, depth ambiguity, and temporal jitter, particularly under extreme viewpoints or complex self-occlusions.

To address these limitations, current research increasingly integrates explicit human priors. While diffusion-based methods~\cite{dreamfusion,zero-1-to-3, dreamhuman,sith,diffhuman,seeavatar} offer impressive texture generation, their iterative denoising prevents real-time application. Alternatively, feed-forward models like HumanRAM~\cite{humanram} bridge parametric body models with neural rendering by conditioning architectures on SMPL-X~\cite{smplx,mon3tr} priors. This highlights the critical role of geometric guidance for handling sparse inputs~\cite{humanram,lvsm}. However, extracting these priors requires a computationally heavy auxiliary regression model at runtime, which blocks real-time performance and occupies more space. Consequently, there remains a critical need for unified frameworks that efficiently utilize robust structural priors to balance rendering fidelity with computational efficiency.

\subsection{Feed-forward NVS and 3D Reconstruction}
Generalizable 3D reconstruction and feed-forward NVS have advanced significantly with the emergence of large reconstruction models (LRMs), which leverage scalable transformers to learn generic priors from extensive datasets~\cite{lrm,lrm-zero}. Unlike traditional optimization-based pipelines, recent foundation models directly map input images to 3D representations or novel views in a feed-forward manner. For 3D reconstruction, methods such as DUSt3R, CUT3R, VGGT~\cite{dust3r,cut3r,ttt3r,human3r,vggt, pi3} have enabled robust pixel-level 3D understanding and direct point cloud regression. Meanwhile, feed-forward NVS frameworks like LVSM and RayZer~\cite{lvsm,rayzer,erayzer} minimize handcrafted inductive biases to deliver remarkable visual quality. Nonetheless, the monolithic self-attention mechanisms employed in these architectures introduce quadratic computational complexity, creating a substantial computational bottleneck for real-time high-resolution applications~\cite{efficient-lvsm}. To address this, recent research has increasingly emphasized architectural efficiency and multi-representation integration. For instance, Efficient-LVSM~\cite{efficient-lvsm} propose decoupled and linearized attention mechanisms to reduce computational overhead while preserving generative quality. Moreover, unified frameworks extend these capabilities by incorporating diverse geometric priors, including depth maps, camera intrinsics, and surface normals, into the reconstruction pipeline~\cite{worldmirror,anysplat,vggt}. These advances suggest that generalizable reconstruction significantly benefits from lightweight multi-representation architectures capable of flexibly adapting to various guidance signals under real-time constraints in practical applications.

\begin{figure}[t!]
  \centering
  \includegraphics[scale=0.57]{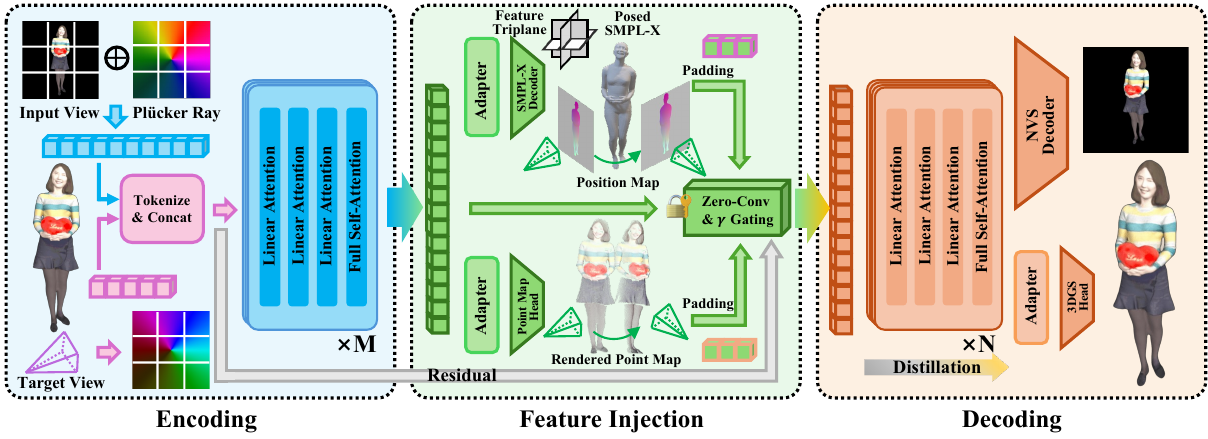} 
  \caption{\textbf{Overview of the PrismMirror architecture.} The framework operates through three cascaded stages: encoding global context, injecting explicit geometric priors (SMPL-X and point clouds), and decoding into NVS or 3DGS (accelerated by a progressive linear attention distillation strategy).}
  \label{fig:pipeline}
  \vspace{-2.3em}
\end{figure}

\subsection{Human Mesh Reconstruction}
Human mesh recovery (HMR) is fundamental for understanding 3D human pose and shape from monocular imagery. The field has transitioned from computationally intensive optimization-based methods, such as SMPLify-X~\cite{smplify-x}, to efficient regression-based frameworks. Early regression approaches, including HMR, SPIN, and PARE~\cite{hmr,spin,pare}, utilize convolutional neural networks (CNNs) to map pixels directly to SMPL body parameters. While effective for global pose estimation, these methods often struggle to capture fine-grained details, leading to misalignments between the projected mesh and the input image~\cite{spin,pare}. The integration of Vision Transformers (ViT) has recently propelled the field toward ``expressive'' whole-body recovery, enabling models to capture intricate facial and hand details with increased accuracy~\cite{hmr,osx}. Current state-of-the-art methods, such as PEAR~\cite{pear}, leverage pixel-level supervision and differentiable rendering to resolve alignment issues and improve fidelity. By estimating parameters for expressive models like SMPL-X~\cite{smplx}, these frameworks provide robust geometric priors that are highly beneficial for guiding complex downstream tasks, including the novel view synthesis and animation targeted by our work~\cite{pear,osx,smplx}.

In contrast to the aforementioned pipelines, PrismMirror positions itself at the intersection of these three domains. By unifying the expressive geometric priors of advanced HMR models into a linearized feed-forward NVS architecture, we bypass the computational bottlenecks of heavy external conditions and quadratic attention, enabling real-time high-fidelity human telepresence.





\section{Method}

In this section, we present PrismMirror, an efficient geometry-guided framework for real-time frontal view synthesis. To resolve the high-frequency artifacts and temporal instability inherent in purely data-driven models without sacrificing inference speed, we propose to learn, decode, and inject explicit 3D human priors directly within the network pipeline. We first present the preliminaries of NVS  (Sec. \ref{sec:preliminary}), and then detail our cascaded architecture, as shown in Fig.~\ref{fig:pipeline}, designed for the generation of structural priors and multi-head decoding (Sec. \ref{sec:architecture}). Finally, we describe the multi-stage learning objectives (Sec. \ref{sec:objective}) and introduce our cascade distillation strategy, which is specifically tailored to overcome the quadratic computational bottlenecks of standard transformers (Sec. \ref{sec:distillation}).

\subsection{Preliminary}
\label{sec:preliminary}

We formulate single-image NVS as a feed-forward sequence-to-sequence spatial mapping problem. Following recent advancements in transformer-based view synthesis~\cite{lvsm, humanram}, our framework avoids compute-intensive 3D representations, such as explicit cost volumes or rigid epipolar geometric projections, during the initial feature extraction phase. Instead, it leverages the global receptive field of transformers, combined with ray-based spatial conditions, to implicitly establish cross-view correspondences.

Given a single source RGB image $\mathbf{I}_s \in \mathbb{R}^{H \times W \times 3}$ with known camera intrinsics and extrinsics, we compute a pixel-wise Plücker ray \cite{plucker} embedding map $\mathbf{P}_s \in \mathbb{R}^{H \times W \times 6}$. The Plücker ray representation is chosen over standard Cartesian coordinates as it inherently avoids origin singularities and preserves 3D line geometry. Specifically, for each pixel, the ray is defined as:
\begin{equation}
    \mathbf{r} = (\mathbf{d}, \mathbf{o} \times \mathbf{d}),
\end{equation}
where $\mathbf{o} \in \mathbb{R}^{3}$ denotes the camera's optical center and $\mathbf{d} \in \mathbb{R}^{3}$ represents the normalized viewing direction.


During the tokenization phase, both the source image and its Plücker ray \cite{plucker} map are divided into non-overlapping local patches of size $p \times p$. A learnable linear projection layer maps these flattened and concatenated patches into a sequence of high-dimensional source tokens $\mathbf{x}_j \in \mathbb{R}^{d}$:
\begin{equation}
    \mathbf{x}_j = \text{Linear}_{inp}\left(\left[\mathbf{I}_{s,j}, \mathbf{P}_{s,j}\right]\right),
\end{equation}
where $[\cdot, \cdot]$ denotes channel-wise concatenation, and $d$ represents the latent token embedding dimension. Similarly, the geometric condition for the desired target view is entirely encapsulated by its Plücker ray \cite{plucker} map $\mathbf{P}_t$, which is projected into a sequence of target query tokens $\mathbf{q}_j \in \mathbb{R}^{d}$:
\begin{equation}
    \mathbf{q}_j = \text{Linear}_{tar}\left(\mathbf{P}_{t,j}\right).
\end{equation}
The core of the framework is a multi-layer transformer model $\mathcal{T}$, which receives the concatenated sequence of source tokens $\mathbf{x} = \{\mathbf{x}_1, \dots, \mathbf{x}_{l_x}\}$ and target tokens $\mathbf{q} = \{\mathbf{q}_1, \dots, \mathbf{q}_{l_q}\}$. Using cascaded multi-head self-attention, the transformer implicitly performs cross-view correspondence matching and geometric reasoning. The network then predicts the contextualized target view tokens $\mathbf{y}_j \in \mathbb{R}^d$:
\begin{equation}
    \mathbf{y}_1, \dots, \mathbf{y}_{l_q} = \mathcal{T}(\mathbf{x}_1, \dots, \mathbf{x}_{l_x}, \mathbf{q}_1, \dots, \mathbf{q}_{l_q}).
\end{equation}
In traditional transformer-based NVS models, the output tokens are passed through a final linear decoder, followed by a Sigmoid activation, to regress the RGB color values of the target patches: $\hat{\mathbf{I}}^t_j = \text{Sigmoid}(\text{Linear}_{out}(\mathbf{y}_j)) \in \mathbb{R}^{3p^2}$. The predicted patches are then reassembled into a 2D spatial grid to construct the synthesized novel view. However, relying solely on implicit regression without explicit structural priors often results in severe 3D inconsistencies. To address this issue, we introduce explicit geometric interventions directly within our cascaded architecture, ensuring structural stability and visual fidelity.

\subsection{Cascaded Model Architecture}
\label{sec:architecture}
While transformer-based NVS demonstrates extraordinary zero-shot generalization across diverse scenes, its minimal 3D inductive bias often leads to physically implausible floating artifacts, limb distortions, and severe temporal inconsistencies in dynamic human-centric scenarios. To address these limitations without sacrificing real-time inference speed, our PrismMirror introduces a sophisticated cascaded architecture. This architecture is logically divided into three progressive functional stages: encoding, feature injection, and decoding.

\noindent \textbf{Stage I: Initial Global Context Encoding.} The input sequence, comprising concatenated source and target tokens $\mathbf{T}_{in} = [\mathbf{x}, \mathbf{q}]$, is first processed in the encoding stage, which consists of $M$ transformer blocks. This stage acts as a robust generalized feature extractor, establishing rudimentary cross-view correlations based purely on data-driven attention mechanisms. The output is an intermediate latent token representation $\mathbf{T}_{mid} \in \mathbb{R}^{(l_{x}+l_{q}) \times d}$, containing rich appearance textures but lacking explicit geometric constraints:
\begin{equation}
  \mathbf{T}_{mid} = \mathcal{T}^{(0 \rightarrow M-1)}\left(\mathbf{T}_{in}\right).
\end{equation}



\noindent \textbf{Stage II: Structural Geometry Prior Parsing and Feature Injection.} Rather than relying on slow external auxiliary networks to extract structural priors during inference, PrismMirror generates these priors directly within the framework. This is achieved through two parallel prediction branches that fork from the intermediate tokens $\mathbf{T}_{mid}$, providing robust structural guidance for subsequent rendering.

The upper semantic branch utilizes a dedicated spatial adapter and an SMPL-X decoder \cite{pear} to predict a complete set of parametric human variables $\mathbf{\Theta}_{smplx} = \{\boldsymbol{\beta}, \boldsymbol{\theta}, \boldsymbol{\psi}, \boldsymbol{\pi}_{cam}\}$, which represent shape, articulation pose, facial 
\begin{wrapfigure}{l}{0.4\textwidth} 
  \centering
    \vspace{-1.3em}
  \includegraphics[width=0.9\linewidth]{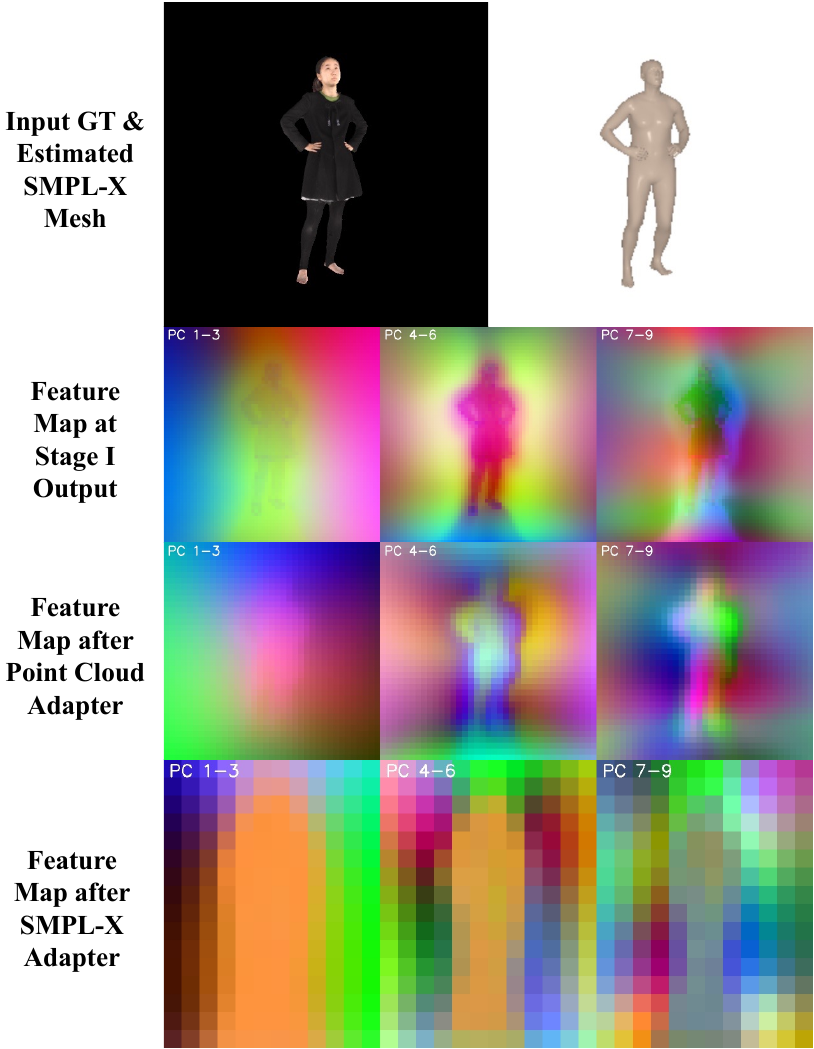} 
  \caption{\textbf{Visualization of geometry feature injection.} By generating explicit spatial priors (point clouds and SMPL-X) inside the model, our architecture effectively anchors pure data-driven features to capture precise body topology and high-frequency details.}
    \vspace{-2.3em}
  \label{fig:featuremap}
\end{wrapfigure}
expression, and orthographic camera projection. Using a differentiable rasterizer, the decoded SMPL-X vertices are projected into the target perspective to generate a dense pixel-aligned position map. To extract high-dimensional semantic representations, this map serves as a dense coordinate scaffold for querying a learned feature triplane, yielding robust pose-aware spatial features $\mathbf{F}_{smpl}$. 

Concurrently, to capture details beyond the bare-body topology provided by SMPL-X, such as clothing and hair, the lower physical branch employs another adapter with a point map head \cite{dpt,worldmirror} to decode $\mathbf{T}_{mid}$ into a rigorous 3D spatial representation. This produces a rendered point map that yields complementary geometric features $\mathbf{F}_{pts}$. The visualization results of the features learned by Stage I are shown in Fig.~\ref{fig:featuremap}.

To integrate these dual geometric features without destabilizing the pre-trained weights of the main backbone, both feature maps undergo spatial padding to align their dimensions and are processed by convolutional layers initialized with weights and biases set to zero, denoted by $\text{Conv}_{zero}(\cdot)$. Crucially, to further preserve low-level spatial frequencies and prevent gradient vanishing, we establish a long-skip residual connection directly from the initial tokenized sequence $\mathbf{T}_{in}$. The dual processed features and the residual \cite{resnet} are combined with the intermediate tokens $\mathbf{T'}_{mid}$, modulated by three distinct learnable gating parameters $\gamma_{res}$, $\gamma_{smpl}$, and $\gamma_{pts}$:
\begin{equation}
    \begin{aligned}
    \mathbf{T'}_{mid} = \mathbf{T}_{mid} &+ \gamma_{smpl} \cdot \text{Flatten}(\text{Conv}_{zero}(\mathbf{F}_{smpl})) \\
    &+ \gamma_{pts} \cdot \text{Flatten}(\text{Conv}_{zero}(\mathbf{F}_{pts})) + \gamma_{res} \cdot \mathbf{T}_{in}.
    \end{aligned}
\end{equation}
This Zero-Conv \& $\gamma$ Gating design \cite{controlnet} ensures that, at the start of training, the model behaves identically to the vanilla pre-trained backbone, gradually learning to incorporate the structural guidance as the gating parameters $\gamma$ update.

\noindent \textbf{Stage III: Dual-Head Representation Decoding.} The geometry-enhanced tokens $\mathbf{T}'_{mid}$ are passed through the decoding stage, which consists of $N$ transformer blocks \cite{lvsm}. This stage contextualizes the injected geometric priors with global appearance features, resulting in the final output tokens $\mathbf{T}_{out}$:
\begin{equation}
    \mathbf{T}_{out} = \mathcal{T}_{dec}^{(0 \rightarrow N-1)}(\mathbf{T'}_{mid}).
\end{equation}
To facilitate comprehensive coarse-to-fine learning, PrismMirror employs a dual-head decoding design rather than relying on a single monolithic output. The final tokens $\mathbf{T}_{out}$ are distributed to two parallel specialized decoding heads:
\begin{equation}
    (\mathbf{\hat{I}}_{t}, \mathbf{\Theta}_{gs}) = (\text{NVS\_Decoder}(\mathbf{T}_{out}), \text{3DGS\_Head}(\mathbf{T}_{out})),
\end{equation}
where the NVS decoder generates the synthesized novel view 
$\mathbf{\hat{I}}_{t}$, leveraging the rich appearance capacity of 2D generative networks, and the 3DGS head outputs volumetric consistency cues $\mathbf{\Theta}_{gs}$, enforcing strict geometric alignment and spatial coherence.
This cascaded dual-head design ensures that the final synthesis benefits from both high-quality texture generation of 2D networks (via the NVS decoder) and the structural consistency of explicit 3D representations (via the 3DGS head adapter) \cite{dpt,worldmirror}.

\subsection{Learning Objective}
\label{sec:objective}

PrismMirror is trained end-to-end using a composite multi-stage loss function to enforce explicit 3D geometry consistency during the intermediate feature injection (Stage II) and photometric fidelity during the final decoding stage (Stage III). The total loss is formulated as:
\begin{equation} \label{total}
    \mathcal{L}_{total} = \lambda_{nvs}\mathcal{L}_{nvs} + \lambda_{gs}\mathcal{L}_{gs} + \lambda_{pts}\mathcal{L}_{pts} + \lambda_{smplx}\mathcal{L}_{smplx}.
\end{equation}

\noindent \textbf{Intermediate Geometric Supervision:}
To guarantee that the generated structural priors possess strict spatial awareness, we apply intermediate supervision to both the point map head \cite{dpt,worldmirror} and the SMPL-X decoder \cite{pear}. For spatial geometry, the point cloud loss $\mathcal{L}_{pts}$ aligns the predicted 3D coordinates $\mathbf{X}$ with the pseudo-ground-truth $\mathbf{X}_{gt}$. To capture multi-frequency structural consistency and accommodate aleatoric uncertainty, we compute a multi-scale gradient loss, modulated by a predicted confidence map $\mathbf{C} \in \mathbb{R}^{H \times W \times 1}$:
\begin{equation}
    \mathcal{L}_{pts} = \sum_{k} \frac{1}{2^{k}} || \mathbf{C} \odot (\nabla_{k}\mathbf{X} - \nabla_{k}\mathbf{X}_{gt}) ||_{1},
\end{equation}
where $\nabla_{k}$ denotes the spatial gradient at decimation scale $k$. This effectively penalizes geometric distortions while preserving sharp boundaries.

Meanwhile, to enforce consistency in the learned geometric priors, the SMPL-X parameters predicted by the model are  constrained using a mean squared error (MSE) against the teacher model's highly accurate SMPL-X parameters:
\begin{equation}
    \mathcal{L}_{smplx} = ||\boldsymbol{\beta}_{s} - \boldsymbol{\beta}_{t}||_{2}^{2} + ||\boldsymbol{\theta}_{s} - \boldsymbol{\theta}_{t}||_{2}^{2} + ||\boldsymbol{\psi}_{s} - \boldsymbol{\psi}_{t}||_{2}^{2} + ||\boldsymbol{\pi}_{s} - \boldsymbol{\pi}_{t}||_{2}^{2},
\end{equation}
where $\boldsymbol{\beta}$, $\boldsymbol{\theta}$, $\boldsymbol{\psi}$, and $\boldsymbol{\pi}$ denote the human body shape, body pose, facial expression, and camera projection parameters, respectively.

\noindent  \textbf{Final Photometric Supervision:}
To ensure high-fidelity visual appearance and authentic identity preservation, the Stage III decoding heads are strictly supervised by the ground-truth target images $\mathbf{I}_{t}$. The synthesized target image $\mathbf{\hat{I}}_{t}$ is optimized using a combination of MSE and the learned perceptual image patch similarity (LPIPS) \cite{lpips} metric. This loss $\mathcal{L}_{nvs}$ captures both pixel-level accuracy and high-frequency perceptual textures:
\begin{equation}
    \mathcal{L}_{nvs} = ||\mathbf{\hat{I}}_{t} - \mathbf{I}_{t}||_{2}^{2} + \lambda_{lpips} \cdot \text{LPIPS}(\mathbf{\hat{I}}_{t}, \mathbf{I}_{t}).
\end{equation}

Similarly, the explicit volumetric output rendered from the 3DGS head \cite{dpt,worldmirror}, denoted by $\mathbf{\hat{I}}_{gs}$, is supervised using an identical metric combination $\mathcal{L}_{gs}$ to enforce robust physical view consistency. We empirically set the balancing coefficients $\lambda_{nvs} = 1.0$, $\lambda_{gs} = 1.0$, $\lambda_{pts} = 100.0$, $\lambda_{smplx} = 1.0$, and $\lambda_{lpips} = 1.0$ during the optimization process.

\begin{wrapfigure}{r}{0.4\textwidth} 
  \centering
  \vspace{-3em}
  \includegraphics[width=\linewidth]{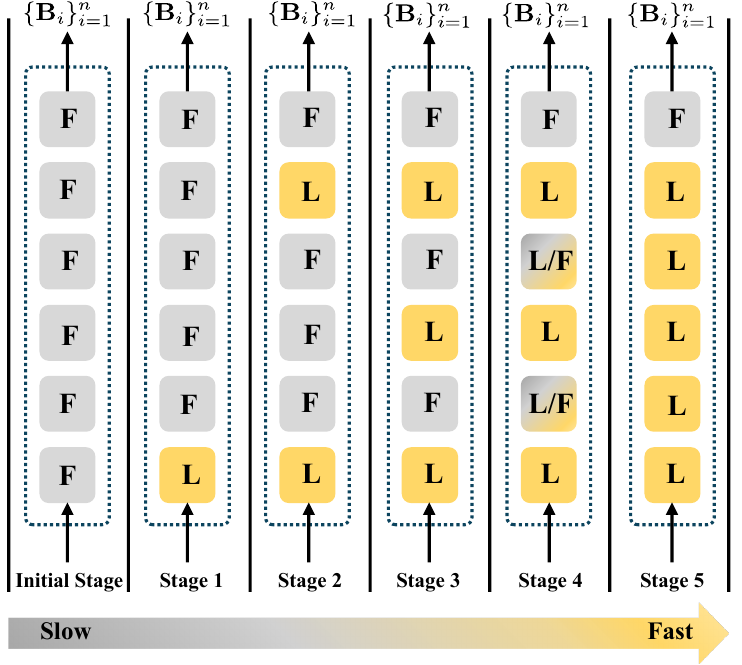} 
  \vspace{-1.2em}
  \caption{\textbf{Progressive distillation.}}
  \label{fig:distillation}
  \vspace{-2.3em}
\end{wrapfigure}

\vspace{-3mm}
\subsection{Model Distillation}
\label{sec:distillation}

Despite the impressive synthesis quality achieved by the base architecture of PrismMirror, the inherent quadratic computational complexity $\mathcal{O}(N^{2})$ of standard multi-head self-attention fundamentally restricts its applicability \cite{transformer,flashattention,flashattention2,fla,sana}. As the input image resolution increases, the token sequence length $N$ grows exponentially, making 24 FPS real-time inference mathematically impossible and imposing severe memory bottlenecks. The standard attention mechanism calculates the interaction matrix directly:
\begin{equation}
    \text{Attention}(\mathbf{Q}, \mathbf{K}, \mathbf{V}) = \text{softmax}\left(\frac{\mathbf{Q}\mathbf{K}^T}{\sqrt{d}}\right)\mathbf{V}.
\end{equation}
To enable real-time performance, we replace standard attention with an advanced linear attention \cite{fla,sana} mechanism, which reduces both time and memory complexity to $\mathcal{O}(N)$. By removing the non-linear softmax operation along the sequence dimension and applying a deterministic feature mapping kernel $\phi(\cdot)$ (e.g., ELU activation plus one) to the Query and Key matrices, we can mathematically exploit the associative property of matrix multiplication:
\begin{equation}
    \text{LinearAttn}(\mathbf{Q}, \mathbf{K}, \mathbf{V}) = \phi(\mathbf{Q})\left(\phi(\mathbf{K})^T \mathbf{V}\right).
\end{equation}

However, directly substituting all attention modules and training from scratch leads to catastrophic forgetting of pre-trained spatial priors. This results in substantial degradation of high-frequency textures and geometric stability. To overcome this challenge, we propose a dynamic linear attention stitching strategy.

As illustrated in Fig.~\ref{fig:distillation}, we implement a progressive group-based distillation protocol. We designate the original fully converged base model that utilizes pure quadratic attention as the frozen expert teacher ($\mathcal{T}_{teacher}$), and our hybrid PrismMirror network as the student ($\mathcal{T}_{student}$). We logically divide the $M+N$ transformer layers into sequential groups, with every 6 attention blocks forming a functional unit.

To ensure a smooth transition, we introduce a temporal unfreezing schedule. Every 5,000 training steps, linear attention \cite{fla} (L) modules are incrementally stitched into the network, replacing full or flash attention \cite{flashattention,flashattention2} (F) modules within each group. Importantly, to preserve the representational capacity and prevent any noticeable performance degradation, the final attention layer within each 6-block group is permanently retained as full attention. This ``majority-linear, minority-full'' hybrid design enables lightweight linear layers to handle localized feature processing, while periodic full attention layers maintain the global receptive field necessary for complex 3D reasoning.


\begin{figure}[tb]
  \centering
  \includegraphics[scale=0.49]{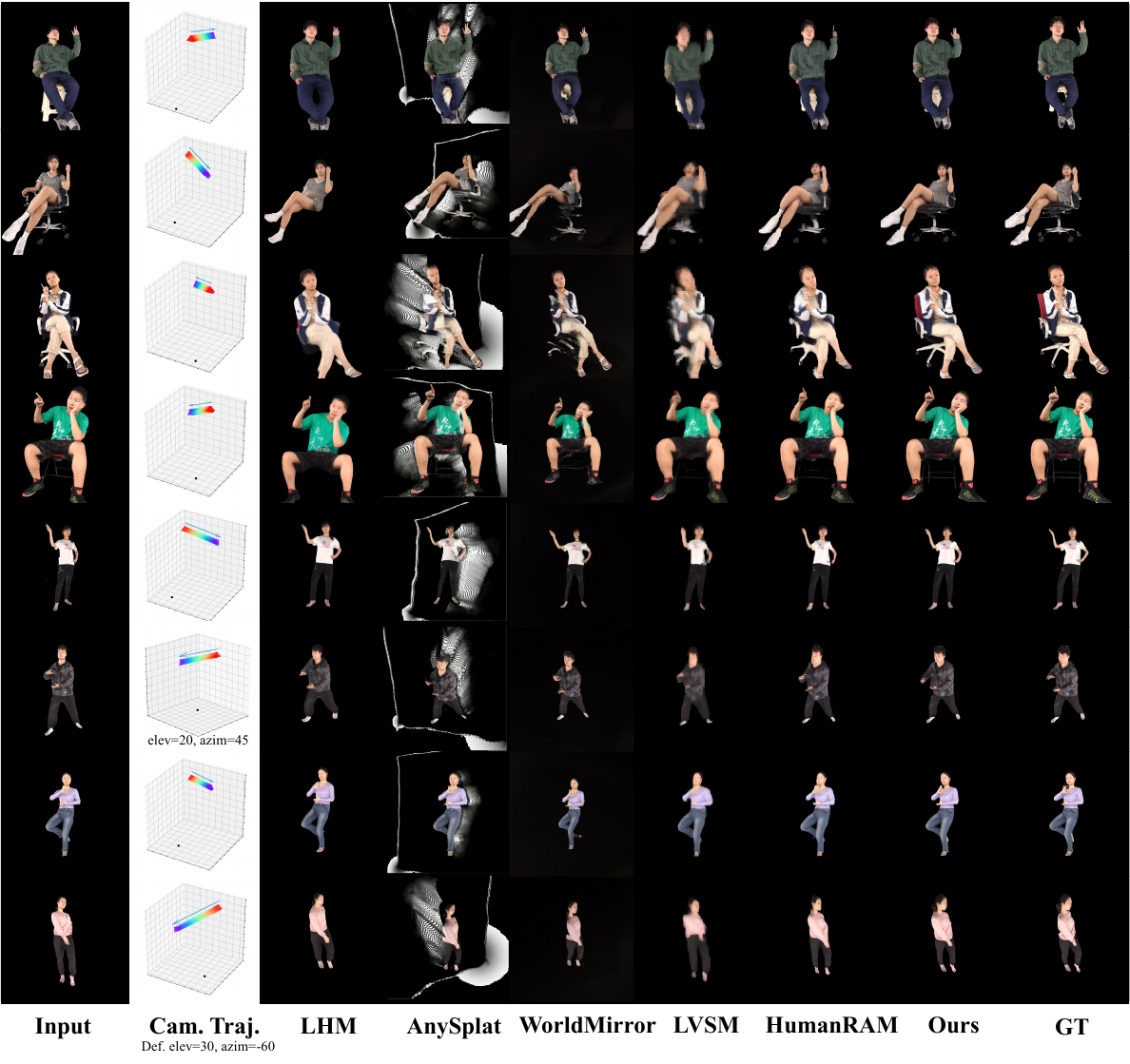} 
  \caption{\textbf{Qualitative comparisons on THuman2.1 and THumanSit.} PrismMirror synthesizes sharper details in high-frequency regions (e.g., faces and hands) compared to baselines, successfully avoiding severe floating artifacts.}
  \label{fig:thuman}
\vspace{-2.7em}
\end{figure}

To enforce representational consistency during this dynamic transition, we compute a pure distillation loss. We extract the intermediate hidden token states from uniformly sampled layers $\mathcal{S}$ in both the frozen teacher and the evolving student, minimizing their MSE loss:
\begin{equation}
    \mathcal{L}_{distill} = \frac{1}{|\mathcal{S}|} \sum_{i \in \mathcal{S}} ||\mathcal{T}_{student}^{(i)} - \mathcal{T}_{teacher}^{(i)}||_{2}^{2}.
\end{equation}
This loss is integrated into the total objective $\mathcal{L}_{total}$ in Eq. \ref{total}. Through this progressive distillation process, our lightweight hybrid student network effectively inherits the rich geometric and photometric knowledge of the quadratic baseline. This enables PrismMirror to achieve robust high-fidelity synthesis at real-time frame rates, decisively breaking the traditional performance barrier .


\begin{table}[t!]
    \centering
    \caption{\textbf{Quantitative comparison on static datasets.} Best results are in \textbf{bold} and second best are \underline{underlined}.}
    \label{tab:static}
    \scriptsize 
    \begin{tabular}{l|ccc|ccc|ccc|c}
        \toprule
        \multirow{2}{*}{\textbf{res=512}} & \multicolumn{3}{c|}{\textbf{THuman2.1}} & \multicolumn{3}{c|}{\textbf{THumanSit}} & \multicolumn{3}{c|}{\textbf{2K2K}} & \multirow{2}{*}{FPS$\uparrow$} \\
        \cmidrule(lr){2-4} \cmidrule(lr){5-7} \cmidrule(lr){8-10}
        & PSNR$\uparrow$ & SSIM$\uparrow$ & LPIPS$\downarrow$ & PSNR$\uparrow$ & SSIM$\uparrow$ & LPIPS$\downarrow$ & PSNR$\uparrow$ & SSIM$\uparrow$ & LPIPS$\downarrow$ & \\
        \midrule
        LHM\cite{lhm}         &  24.75    &   0.9357   &   0.0623    &    22.80   &   \underline{0.9081}   &   0.0913    &   21.06    &   0.8767   &    0.0943   & 0.66 \\
        AnySplat\cite{anysplat}  & 15.44 &    0.7365   &   0.2704   &    12.98   &    0.6141   &   0.3605   &    14.50   &    0.6847   &   0.2860   &     4.66   \\
        WorldMirror\cite{worldmirror} &   23.92    &   0.3289   &    0.0853   &     19.76  &   0.3008   &    0.1549   &   20.81    &   0.3182   &    0.0986   & \underline{6.18}\\
        LVSM\cite{lvsm}        &    26.46   &   0.9452   &   0.0603    &  22.68     &   0.8851   &   0.1208    &    21.78   &  0.8897    &    0.0922   & 3.10\\
        HumanRAM\cite{humanram}    &   \underline{27.96}    &  \underline{0.9551}    &    \underline{0.0371}   &   \underline{23.49}    &    0.9010  &    \underline{0.0814}   &    \underline{26.81}   &   \underline{0.9369}   &   \textbf{0.0343}    &  4.84 \\
        \midrule
        \textbf{Ours} & \textbf{31.23} &   \textbf{0.9670}   &    \textbf{0.0280}   &  \textbf{26.88}    &   \textbf{0.9300}   &   \textbf{0.0563}    &   \textbf{27.57} & \textbf{0.9382} & \underline{0.0375}    & \textbf{24.07}\\
        \bottomrule
    \end{tabular}
\vspace{-2.3em}
\end{table}

\section{Experiment}
\label{sec:blind}

\subsection{Experimental Setup}

\begin{wrapfigure}{r}{0.4\textwidth} 
  \centering
  \vspace{-2.3em}
  \includegraphics[width=\linewidth]{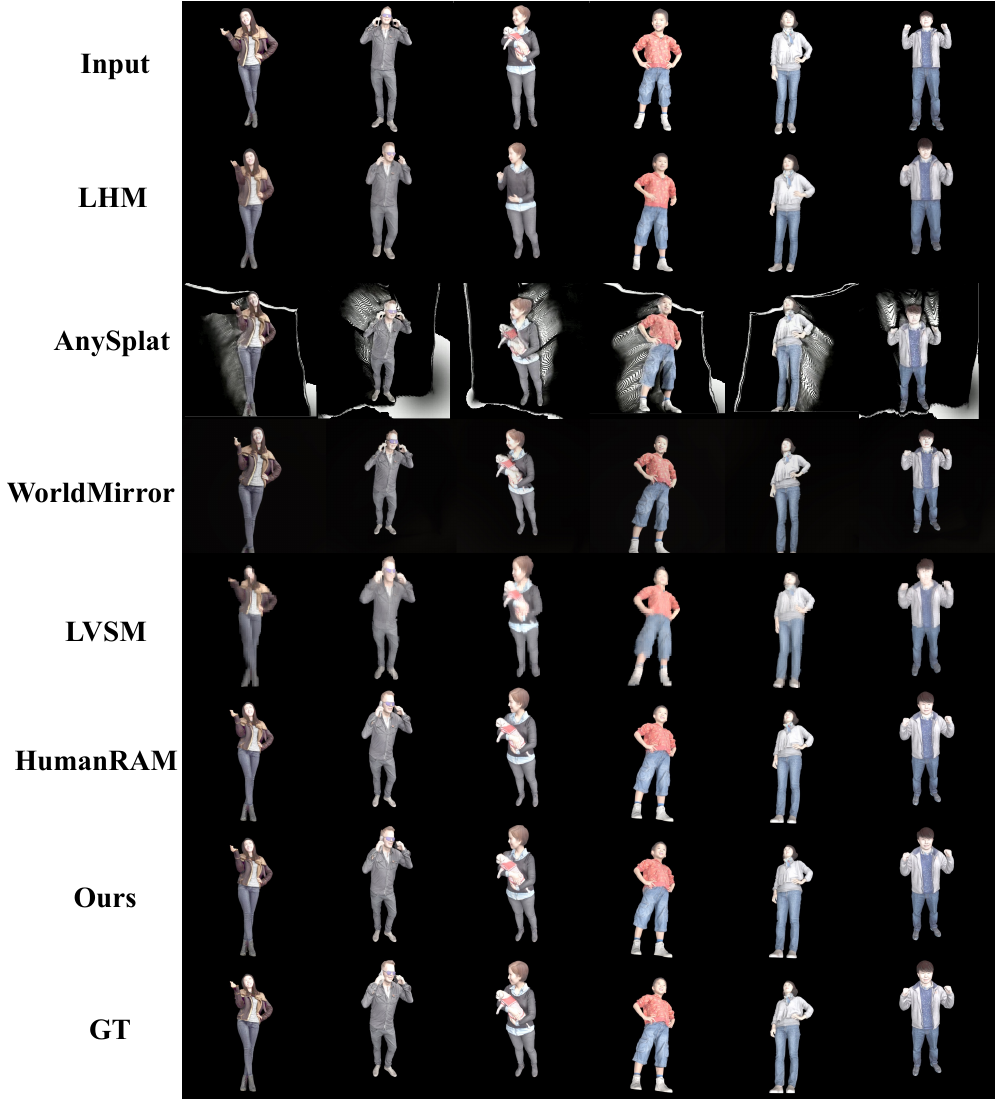} 
  \vspace{-1em}
  \caption{\textbf{Qualitative comparisons on 2K2K.}}
  \label{fig:2k2k}
  \vspace{-2.0em}
\end{wrapfigure}


\textbf{Datasets and Metrics.} We evaluate PrismMirror on static, dynamic, and cross-domain datasets. For static scenes, we render training data using THuman2.1 \cite{thuman2} (2,445 identities), THuman3.0 \cite{thuman3} (20 human-garment combinations), THumanSit \cite{thumansit} (4,700 interacting poses), and 2K2K \cite{2k2k} (2,050 poses). We sample 100 random views within a 30-degree, 2-meter spherical region around a central view for training. The test set includes 50 uniformly sampled identities from THuman2.1, THuman3.0, and 2K2K. For dynamic scenes, we use 6 identities from ActorsHQ \cite{actorshq}, splitting frames in a 4:1 ratio for fine-tuning and testing. We select 13 specific cameras within a 30-degree range for supervision, using camera 127 as the frontal input. For cross-domain zero-shot evaluation, we extract ~2,000 frames from MVHumanNet \cite{mvhumannet} (Part 1), randomly selecting a middle-level camera as input and a neighboring camera within a 30-degree range as the target. To measure visual fidelity, we report peak signal-to-noise ratio (PSNR), structural similarity (SSIM), and LPIPS \cite{lpips}, compared with diverse baselines, avatar-based models (LHM \cite{lhm}), feedforward reconstruction models (AnySplat \cite{anysplat} and WorldMirror \cite{worldmirror}) and NVS models (LVSM \cite{lvsm} and HumanRAM \cite{humanram}). To assess computational efficiency, we measure inference frames per second (FPS).

\textbf{Implementation Details.} The full-attention teacher model is trained on eight A6000 GPUs (batch size 32) for 20,000 steps. We then distill it into the linear-attention student model on a single A6000 GPU (batch size 4) for 6,000 steps. Fine-tuning on ActorsHQ is conducted for 6,000 steps using a single GPU. Linear attention is implemented based on SANA \cite{sana}, with layer parameters $M=12$ and $N=24$. All FPS tests are conducted on a single A6000 GPU.

\begin{figure}[tb]
  \centering
  \includegraphics[scale=0.37]{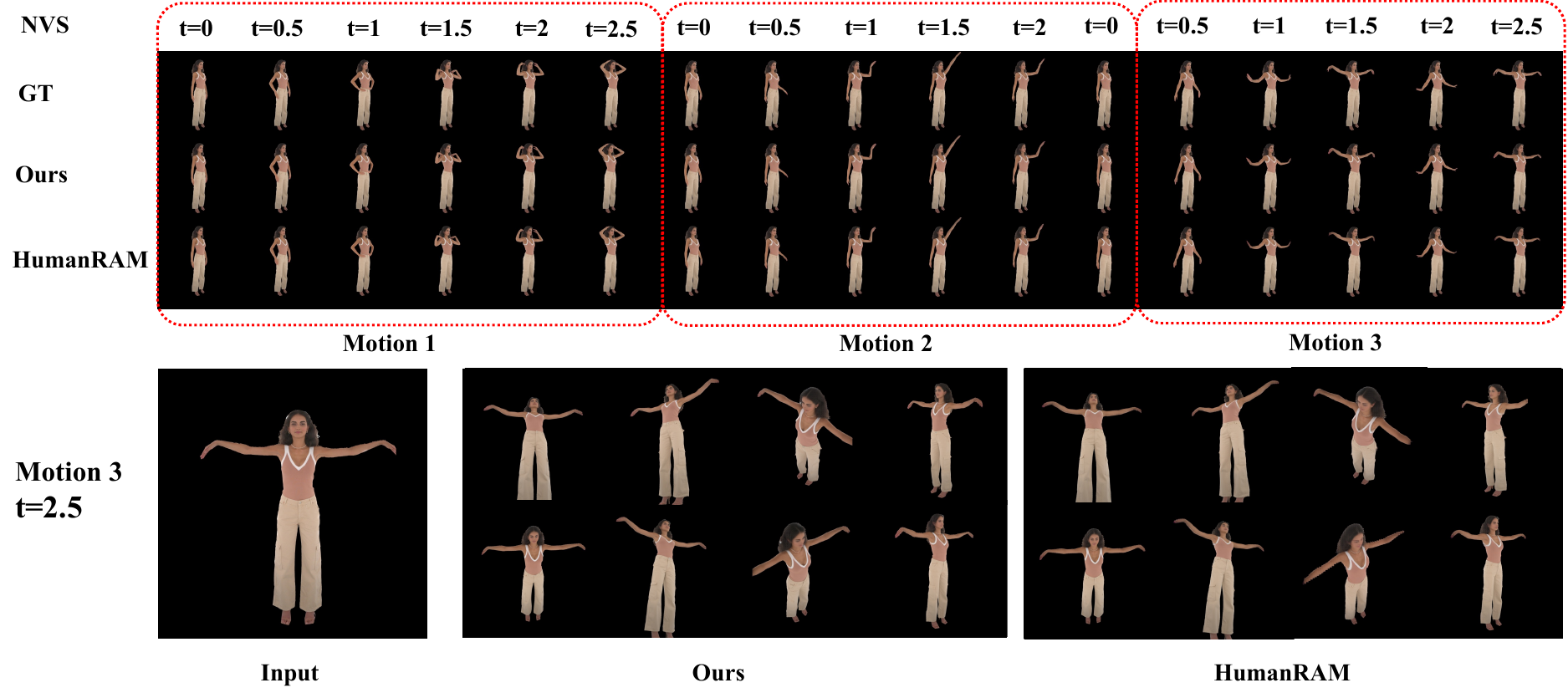} 
  \vspace{-1em}
  \caption{\textbf{Spatiotemporal consistency evaluation on ActorsHQ.} Unlike baselines that suffer from temporal flickering, our explicit geometric anchors ensure highly stable rendering across continuous motion sequences.}
  \label{fig:actorshq}
\vspace{-2.0em}
\end{figure}

\subsection{Comparison}


\begin{wraptable}{r}{0.5\linewidth} 
    \centering
    \vspace{-3.5em}
    \caption{\textbf{Comparison on video datasets.} Best results are in \textbf{bold}.}
    \label{tab:dynamic}
    \scriptsize
    \begin{tabular}{l|ccc}
        \toprule
        \multirow{2}{*}{\textbf{res=512}} & \multicolumn{3}{c}{\textbf{ActorsHQ}}\\
        \cmidrule(lr){2-4}
        & PSNR$\uparrow$ & SSIM$\uparrow$ & LPIPS$\downarrow$ \\
        \midrule
        WorldMirror\cite{worldmirror} & 21.91 & 0.3944 & 0.0980 \\
        HumanRAM\cite{humanram}    & 27.52 & 0.9466 & 0.0380 \\
        \midrule
        \textbf{Ours} & \textbf{29.62} & \textbf{0.9570} & \textbf{0.0307} \\
        \bottomrule
    \end{tabular}
\vspace{-2.3em}
\end{wraptable}



As shown in Table~\ref{tab:static}, PrismMirror significantly outperforms baselines on static datasets, achieving PSNRs of 31.23, 26.88, and 27.57 dB on THuman2.1, THumanSit, and 2K2K (vs. HumanRAM's 27.96, 23.49, and 26.81 dB). Our model also maintains SOTA or near-SOTA SSIM and LPIPS. This advantage stems from our dual-geometry prior injection, which anchors the feature space to deterministic topologies, avoiding depth ambiguity and allocating capacity to high-frequency texture synthesis. Moreover, linear attention boosts inference speed to 24.07 FPS (nearly 5× faster than HumanRAM). Qualitatively (Figs.~\ref{fig:thuman} and~\ref{fig:2k2k}), baselines suffer from severe texture blurring and floating artifacts, especially on faces and hands. 

\begin{wrapfigure}{r}{0.5\textwidth} 
  \centering
  \vspace{-1.3em}
  \footnotesize
  \includegraphics[width=0.95\linewidth]{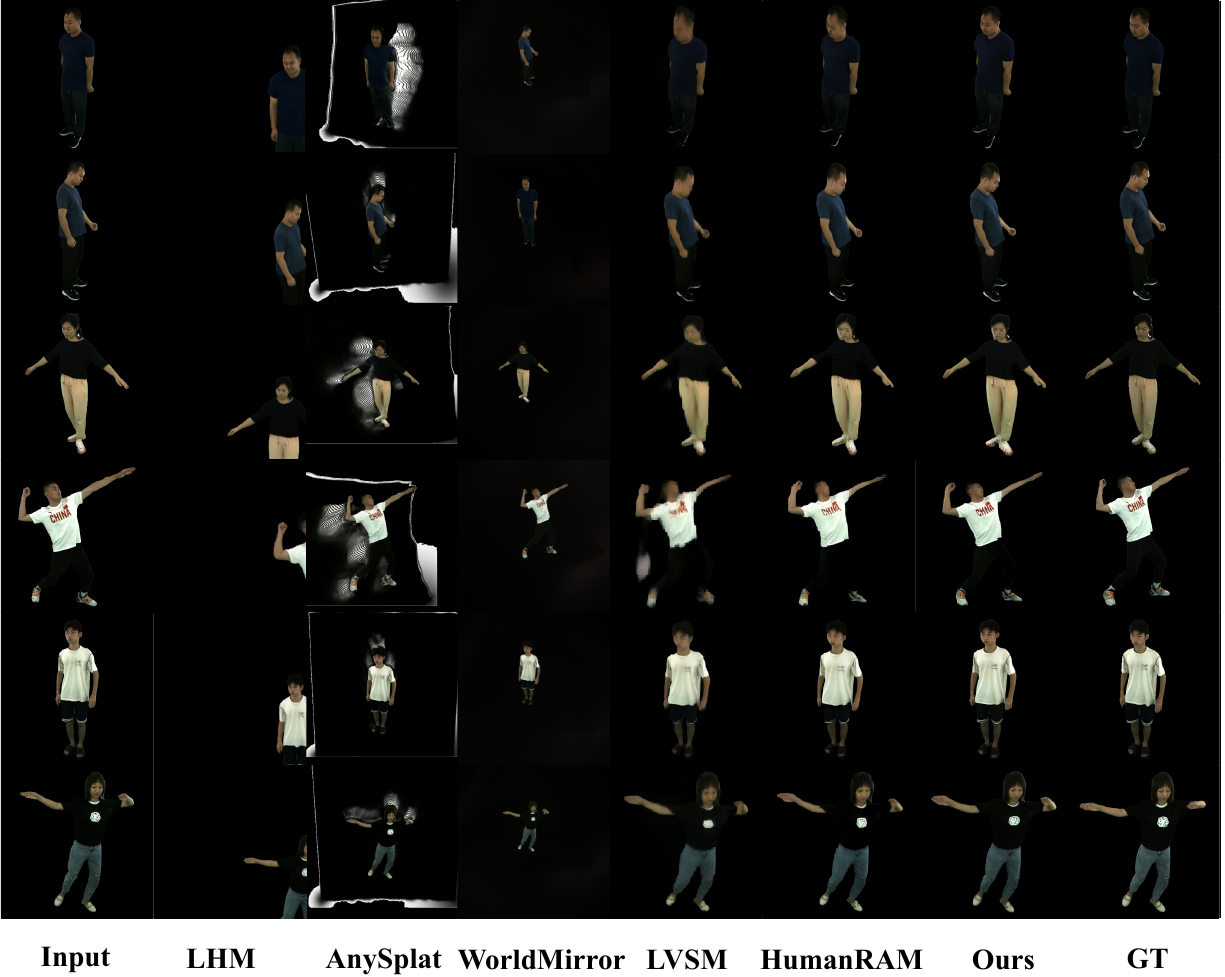} 
  \vspace{-1em}
  \caption{\textbf{Zero-shot cross-domain evaluation MVHumanNet.} PrismMirror demonstrates exceptional identity preservation and structural fidelity on unseen data with non-training camera settings.}
  \label{fig:mvhumannet}
  \vspace{-2.3em}
\end{wrapfigure}

For dynamic scenes shown in Table~\ref{tab:dynamic}, PrismMirror achieves a PSNR of 29.62 dB on ActorsHQ, which is competitive with HumanRAM and WorldMirror. However, as shown in Fig.~\ref{fig:actorshq}, baseline methods suffer from morphological distortions and flickering. In contrast, PrismMirror’s explicit geometric priors ensure strong spatiotemporal coherence. Finally, in cross-domain zero-shot evaluation on MVHumanNet, illustrated in Table~\ref{tab:cross_domain} and Fig.~\ref{fig:mvhumannet}, PrismMirror achieves a PSNR of 25.30 dB, demonstrating robust identity preservation on unseen data.

\subsection{Ablation Study}



\begin{table}[t!]
    \centering
    \scriptsize
    
    \begin{minipage}[t]{0.42\linewidth}
        \centering
        \caption{\textbf{Comparison on cross-domain datasets.} }
        \label{tab:cross_domain}
        \begin{tabular}{l|ccc}
            \toprule
            \multirow{2}{*}{\textbf{res=512}} & \multicolumn{3}{c}{\textbf{MVHumanNet}}\\
            \cmidrule(lr){2-4}
            & PSNR$\uparrow$ & SSIM$\uparrow$ & LPIPS$\downarrow$ \\
            \midrule
            LHM\cite{lhm}         & 21.95 & \textbf{0.9359} & \textbf{0.0789} \\
            AnySplat\cite{anysplat}    & 15.30 & 0.7598 & 0.2596 \\
            WorldMirror\cite{worldmirror} & 22.03 & 0.2256 & 0.1116 \\
            LVSM\cite{lvsm}        & \underline{22.41} & 0.8779 & 0.1277 \\
            HumanRAM\cite{humanram}    & 22.34 & 0.8851 & 0.0971 \\
            \midrule
            \textbf{Ours} & \textbf{25.30} & \underline{0.9029} & \underline{0.0895} \\
            \bottomrule
        \end{tabular}
    \end{minipage}\hfill
    \begin{minipage}[t]{0.54\linewidth}
        \centering
        \caption{\textbf{Ablation Study.} distil. and feat. are short for distillation and feature, respectively.}
        \label{tab:ablation_study}
        \begin{tabular}{l|ccc|c}
            \toprule
            \multirow{2}{*}{\textbf{res=512}} & \multicolumn{3}{c|}{\textbf{Ablation Study}}& \multirow{2}{*}{FPS$\uparrow$}\\
            \cmidrule(lr){2-4}
            & PSNR$\uparrow$ & SSIM$\uparrow$ & LPIPS$\downarrow$ &  \\
            \midrule
            fully linear attention    & 25.90 & 0.9167 & 0.0815 & \textbf{25.59} \\
            w/o progressive distil.    & 22.94 & 0.8758 & 0.1321 & -- \\
            teacher model  & \underline{27.56} & 0.9376 & 0.0384 & 8.77 \\
            w/o pos. map feat.  & 26.43 & 0.9320 & 0.0387 & -- \\
            w/o point map feat. & 27.04 & \textbf{0.9388} & \underline{0.0375} & -- \\
            \midrule
            \textbf{Full Model} & \textbf{27.57} & \underline{0.9382} & \textbf{0.0360} & \underline{24.07} \\
            \bottomrule
        \end{tabular}
    \end{minipage}
    \vspace{-2em}
\end{table}

To validate our architectural decisions, we conduct ablation studies in Table~\ref{tab:ablation_study}. For the dual-geometry injection, removing the position map reduces the PSNR from 27.57 dB to 26.43 dB, as it provides foundational body topology. In contrast, removing the point map reduces the PSNR to 27.04 dB, as it captures deviations like loose clothing. Without these geometric priors, the network relies on implicit estimations, leading to significant degradation in rendering sharpness.

Moreover, the progressive distillation strategy is critical for balancing speed and quality. Removing it entirely causes catastrophic forgetting of spatial priors, drastically reducing the PSNR to 22.94 dB. A fully linear attention configuration (replacing all $M+N$ layers) marginally increases speed to 25.59 FPS but drops the PSNR to 25.90 dB due to limited representational capacity. Conversely, the full-attention teacher model yields a high PSNR of 27.56 dB but a sluggish 8.77 FPS. Our progressive protocol optimally bridges this gap, uniquely achieving a 27.57 dB PSNR and real-time inference at 24.07 FPS.

\section{Conclusion}

In this paper, we introduce PrismMirror, a novel geometry-guided framework for real-time high-fidelity human frontal view synthesis from a single image. By strategically restricting the synthesis to a practical frontal viewing frustum and decoding explicit structural priors from the backbone, e.g., SMPL-X meshes and 3D point clouds, our approach effectively eliminates the reliance on computationally intensive external conditions during inference. To further enhance efficiency, we propose a progressive cascade distillation strategy that dynamically transitions the network from quadratic full attention to lightweight linear attention. This key innovation addresses the traditional tradeoff between structural accuracy and inference speed, enabling PrismMirror to achieve state-of-the-art visual authenticity and spatiotemporal consistency while delivering real-time performance at 24 FPS. Ultimately, PrismMirror paves the way for accessible immersive 3D telepresence without the need for complex multi-camera setups.

\bibliographystyle{splncs04}
\bibliography{main}

\clearpage

\title{Supplementary Material \\[0.5cm]
\large \mdseries \textit{Real-Time Human Frontal View Synthesis from a Single Image}} 

\titlerunning{PrismMirror}
\authorrunning{F. Lin~Y. Hu et al.}

\author{\vspace{-2pt}}

\institute{\vspace{-2pt}}

\maketitle
\vspace{-10mm}
The supplementary material provides limitations and future work in Sec. \ref{sec:limitation}, more implementation details in Sec. \ref{sec:implementation}, more experimental results in Sec. \ref{sec:results}, and some failure cases in Sec. \ref{sec:failure}.

\appendix
\renewcommand{\theHsection}{\Alph{section}}
\section{Limitations and Future Work}
\label{sec:limitation}
While PrismMirror demonstrates strong real-time performance within a frontal viewing frustum of ±30°, its novel view synthesis capabilities remain constrained to this specific range. When the observation angle exceeds this boundary, the model's ability to infer heavily occluded regions degrades, limiting its effectiveness at extreme viewpoints. In the future, we plan to enhance the training pipeline by using more diverse and large-scale multi-view datasets \citesupp{mvhumannet_supp,dnarendering_supp}. This will allow us to expand the model's synthesis capabilities beyond the current frontal constraints, leading to more flexible and comprehensive view synthesis. In addition, our framework currently struggles to generalize in scenarios with large or sudden movements of subjects. If the motion is excessively abrupt or expansive, the model occasionally produces structural artifacts, such as disappearing hands or distorted facial features. To address this issue, we will incorporate stronger and more explicit 3D human spatial priors \citesupp{sapiens_supp}. These priors will enhance the network's understanding of intricate geometric details and biomechanical constraints, ensuring greater visual and structural stability under complex motion conditions.

\section{More Implementation Details}
\label{sec:implementation}

\subsection{Static Dataset Construction}

To construct the static training dataset, we employ a customized rendering pipeline based on Taichi \citesupp{taichi_supp}. For each 3D mesh, we normalize the human height to approximately 1.80 m (with a minor random perturbation of $\pm0.05$ m) and apply random horizontal and depth translations (up to $\pm0.1$ m) to increase spatial diversity. To ensure consistent frontal alignment, we utilize the SMPL-X \citesupp{smplx_supp} orientation parameters to correct the global facing direction. For each subject, we render a total of 105 views: 1 strict frontal center view ($0^\circ$ yaw, $0^\circ$ pitch), 100 randomly sampled views that are uniformly distributed within a viewing frustum of $[-30^\circ, 30^\circ]$ for both yaw and pitch, and 4 extreme corner views ($\pm30^\circ, \pm30^\circ$). The rendering distance is fixed at a radius of 2.0 m. For evaluation, we uniformly sample 50 identities from THuman2.1 \citesupp{thuman2_supp}, THumanSit \citesupp{thumansit_supp}, and 2K2K \citesupp{2k2k_supp}, utilizing all 100 sampled views per identity for testing, while the remaining subjects are allocated strictly to the training set.

\subsection{Dynamic Dataset Configuration}

For dynamic evaluations on ActorsHQ \citesupp{actorshq_supp}, we account for the significant differences in camera coordinate systems by fine-tuning the model directly on this dataset. We select Sequence 1 from 6 identities (Actor01, Actor02, Actor04, Actor06, Actor07, and Actor08). From the approximately 2,500 frames per sequence, we allocate 2,000 frames for training and the remainder for evaluation. We designate Camera 127, which is positioned strictly in the frontal viewing direction, as the fixed input source view. During training, target supervision is sampled exclusively from cameras whose viewing angles fall within a 30° relative offset from Camera 127.

\subsection{Cross-Domain and In-the-Wild Evaluation}

To evaluate cross-domain zero-shot generalization, we randomly extract 2,000 frames from MVHumanNet \citesupp{mvhumannet_supp} (Part 1), which encompasses over 10 identities and 30 diverse appearances. Since the MVHumanNet camera rig is stratified into three elevation layers, we select the middle-tier cameras as the input source and randomly sample adjacent cameras within a 30° range as target views. For in-the-wild testing, we use high-resolution internet images \citesupp{synchuman_supp} or standard smartphone photographs, assuming a default frontal input view. We compute the foundational geometric constraints, such as the Plücker ray \citesupp{plucker_supp} maps, by directly applying the camera intrinsic and extrinsic parameters derived from our static scene rendering pipeline.

\subsection{Training Details}

The training process is divided into two primary stages to ensure effective feature extraction and real-time efficiency:

\begin{itemize}
    \item \textbf{Teacher Pre-training:} We initially train a teacher model with full attention to acquire strong geometric priors. This stage is conducted on 8 NVIDIA RTX A6000 GPUs for 20,000 steps with a batch size of 32. The framework is optimized using the AdamW optimizer with a weight decay of 1e-5. We apply decoupled learning rates: 1e-5 for the main LVSM \citesupp{lvsm_supp} backbone and 2e-4 for the newly initialized dual-prior injection heads. We employ a Cosine Annealing learning rate scheduler, with a minimum learning rate of 1e-6. 

    \item \textbf{Student Distillation:} The pre-trained teacher model is distilled into a lightweight student model over 40,000 steps on a single NVIDIA RTX A6000 GPU, with a batch size of 4. We use a fixed learning rate of 1e-6 for the distillation. To ensure representational consistency, an explicit MSE distillation loss $\mathcal{L}_{distill}$ is computed across the hidden states of evenly spaced layers between the frozen teacher and student models. Specifically, we employ a progressive linear attention stitching strategy: starting at step 5,000, standard attention modules are periodically replaced with linear attention equivalents every 5,000 steps. By step 35,000, 30 out of the 36 transformer blocks are converted to linear attention, while full attention is retained at fixed intervals (e.g., every 6 blocks) to maintain global receptive capacity. Finally, the distilled student model is fine-tuned on the ActorsHQ \citesupp{actorshq_supp} dataset for 6,000 steps on a single GPU.
\end{itemize}

For both the pre-training and distillation stages, we incorporate dual geometric priors, i.e., SMPL-X \citesupp{smplx_supp} position maps and depth-guided point clouds, to improve structural and visual accuracy. Position maps are rendered using a differentiable rasterizer, NVDiffRasterizer~\citesupp{diffrast_supp}, and decoded via tri-plane neural textures with a feature dimension of 32, while point clouds are processed using a Gaussian splatting renderer~\citesupp{3dgs_supp} to impose explicit physical constraints. These features are seamlessly integrated into the backbone using Zero-Conv \citesupp{controlnet_supp} gating mechanisms, parameterized by $\gamma$. This integration ensures that both structural and textural information is effectively utilized, enhancing the model’s ability to synthesize visually consistent and geometrically accurate outputs.


\section{More Experimental Results}
\label{sec:results}
\begin{figure}[b!]
  \centering
  \includegraphics[scale=0.52]{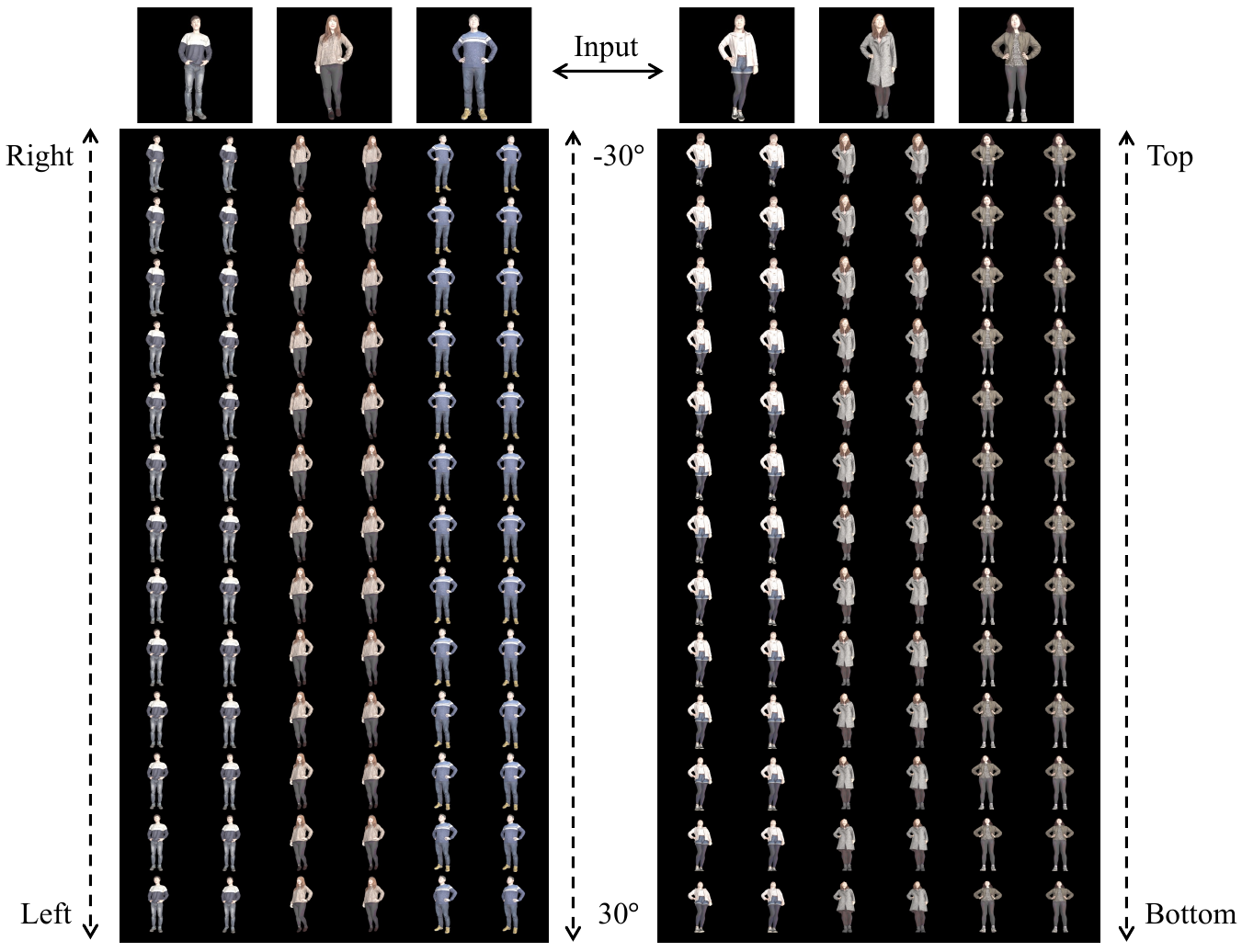} 
  \caption{\textbf{Evaluation of spatial consistency in different directions.} \textbf{Left:} Results for 3 IDs during horizontal camera rotation, moving from right to left (from $-30^{\circ}$ to $30^{\circ}$). \textbf{Right:} Results for 3 IDs during vertical camera rotation, moving from top to bottom (from $-30^{\circ}$ to $30^{\circ}$). In both sets, images 
  are rendered at $5^{\circ}$ intervals. PrismMirror synthesizes sharper details in high-frequency regions, such as faces and hands, compared to baselines, successfully avoiding severe floating artifacts.}
  \label{fig:spatial}
\end{figure}

\begin{figure}[tb]
  \centering
  \includegraphics[scale=0.63]{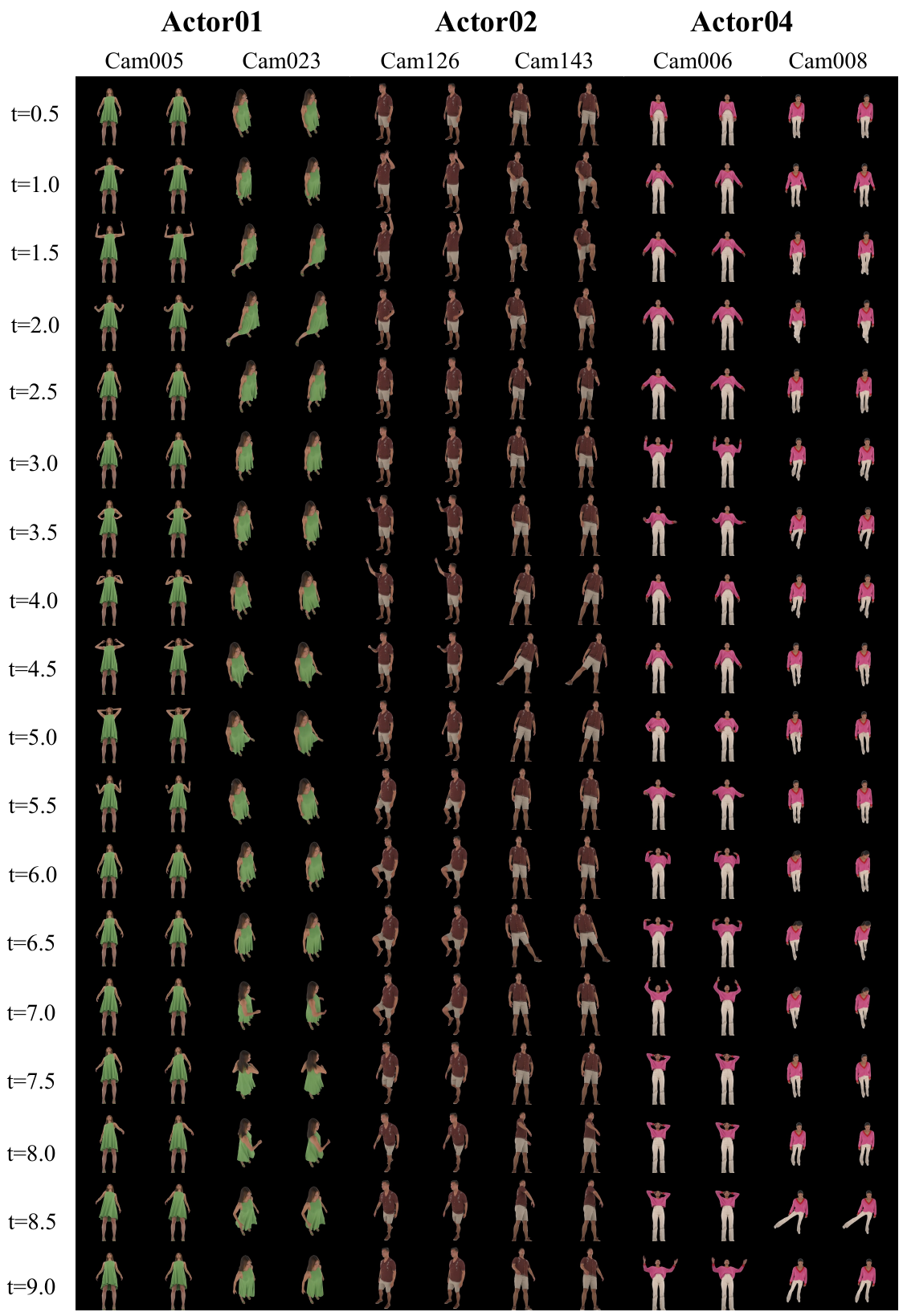} 
  \caption{\textbf{Evaluation of temporal consistency on ActorsHQ.} Here, we present long-sequence results for Actor01, Actor02, and Actor04 from different viewpoints. Even with significant poses, our method still maintains temporal consistency and produces reasonable NVS results.}
  \label{fig:temporal}
\end{figure}

We provide additional evaluation results on spatiotemporal consistency and in-the-wild testing to further validate the robustness and generalization capability of PrismMirror. As illustrated in Fig. \ref{fig:spatial}, we evaluate spatial consistency under both horizontal and vertical camera rotations, ranging from -30° to 30° at 5° intervals. Our method consistently synthesizes sharp details in high-frequency regions, effectively avoiding the floating artifacts common in baseline methods. Moreover, Fig. \ref{fig:temporal} demonstrates long-sequence temporal consistency on the ActorsHQ \citesupp{actorshq_supp} dataset, specifically featuring Actor01, Actor02, and Actor04 across different viewpoints. Even when subjects perform large-scale complex poses, our approach maintains highly stable rendering and temporally coherent novel views. 

\begin{figure}[tb]
  \centering
  \includegraphics[scale=0.4]{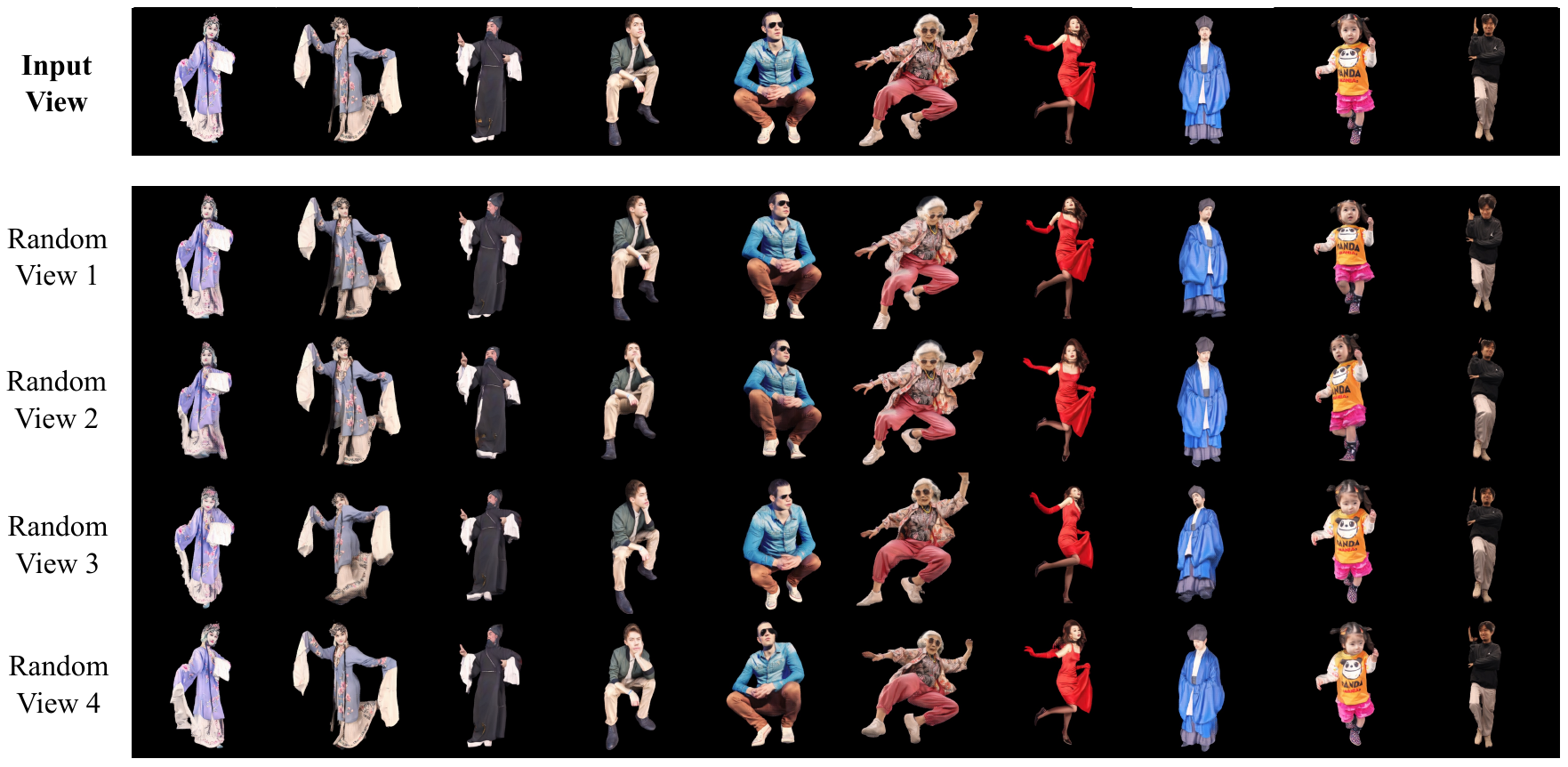} 
  \caption{\textbf{Zero-shot evaluation results on in-the-wild cases.} To validate the practical applicability of our method, we conduct evaluations using in-the-wild data, including Internet images and real-world photographs. The visually appealing results demonstrate the strong generalization capability and robustness of our approach in unseen and complex scenarios.}
  \label{fig:in-the-wild}
\end{figure}

\begin{figure}[tb]
  \centering
  \includegraphics[scale=0.5]{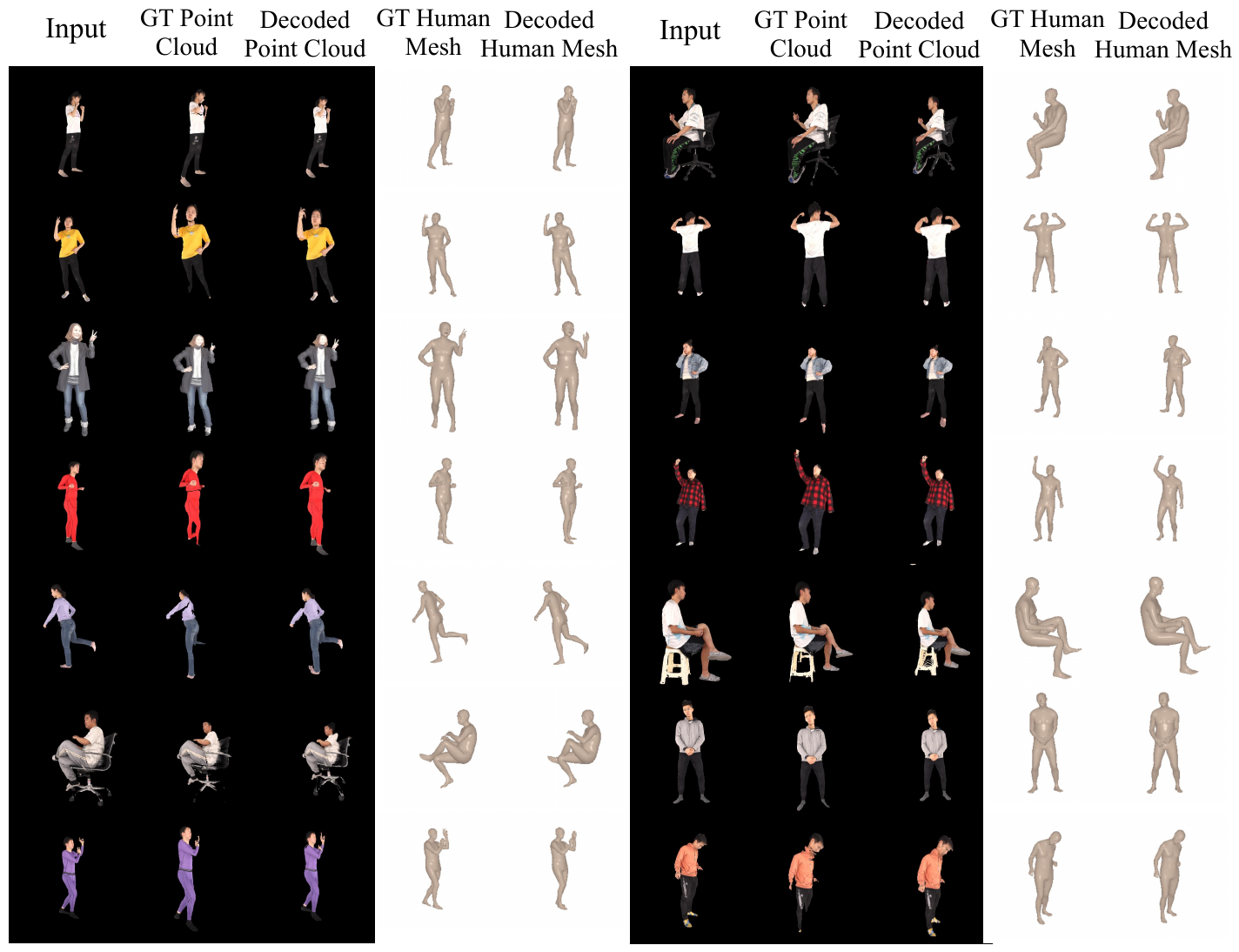} 
  \caption{\textbf{Point cloud and human mesh from the encoding stage.} This figure illustrates the structural learning capability of our 12-layer transformer encoder. By comparing the predicted point clouds and SMPL-X meshes with their respective ground truth (GT), we observe a high degree of alignment. This indicates that our model effectively extracts precise structural information, ensuring strong geometric priors for the subsequent feature injection stage.}
  \label{fig:mesh-and-pc}
\end{figure}

To demonstrate practical applicability, Fig. \ref{fig:in-the-wild} presents zero-shot evaluations on unconstrained in-the-wild data, including diverse Internet images and real-world photographs. The visually appealing results highlight the strong generalization capability of our framework in unseen and complex scenarios. Finally, Fig. \ref{fig:mesh-and-pc} illustrates the structural learning capability of our 12-layer transformer encoder. The predicted point clouds and SMPL-X \citesupp{smplx_supp} meshes closely align with the ground truth, confirming that the initial encoding stage successfully extracts precise structural information to serve as a strong geometric prior for the subsequent feature injection stage.

\section{Failure Cases}
\label{sec:failure}
\begin{figure}[tb]
  \centering
  \includegraphics[scale=0.6]{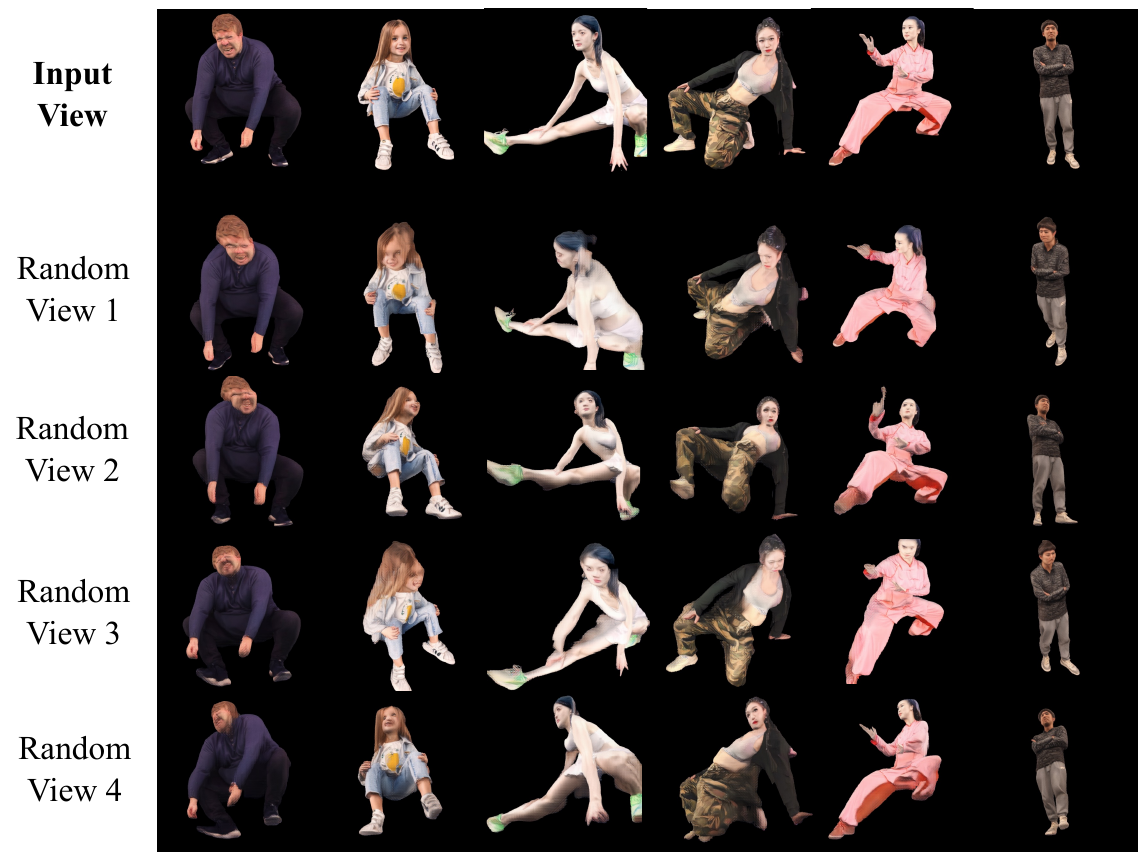} 
  \caption{\textbf{Failure cases.} We observe severe distortions in scenarios involving extreme poses or when the subject is in close proximity to the camera. These artifacts are mainly due to OOD issues, which arise from the limited diversity and coverage of such extreme cases in our current training dataset.}
  \label{fig:failure-cases}
\end{figure}

While PrismMirror achieves real-time high-fidelity synthesis within its targeted domain, it occasionally encounters limitations under highly challenging conditions. As shown in Fig. \ref{fig:failure-cases}, we observe severe rendering distortions in scenarios involving extreme body poses or when the subject is in close proximity to the camera. We attribute these artifacts primarily to out-of-distribution (OOD) issues. The current training dataset lacks sufficient diversity and coverage for such extreme viewpoint and motion variations. To address these failure cases, future improvements will focus on expanding the richness of training data and incorporating more robust biomechanical constraints.

\clearpage

\setcounter{enumiv}{0}
\bibliographystylesupp{splncs04}
\bibliographysupp{supple} 

\end{document}